\documentclass[sigconf, superscriptaddress]{acmart}

\AtBeginDocument{%
  \providecommand\BibTeX{{%
    \normalfont B\kern-0.5em{\scshape i\kern-0.25em b}\kern-0.8em\TeX}}}

\copyrightyear{2023}
\acmYear{2023}
\setcopyright{acmcopyright}\acmConference[WSDM '23]{Proceedings of the Sixteenth ACM International Conference on Web Search and Data Mining}{February 27-March 3, 2023}{Singapore, Singapore}
\acmBooktitle{Proceedings of the Sixteenth ACM International Conference on Web Search and Data Mining (WSDM '23), February 27-March 3, 2023, Singapore, Singapore}
\acmPrice{15.00}
\acmDOI{10.1145/3539597.3570420}
\acmISBN{978-1-4503-9407-9/23/02}

\usepackage{enumitem}
\usepackage{bm}
\usepackage{algorithm}
\usepackage{algpseudocode}
\usepackage{amsmath}
\usepackage{subfigure}
\usepackage{diagbox}
\usepackage{makecell}
\usepackage{multirow}
\usepackage{url}
\usepackage{color, xcolor}
\usepackage{balance}

\usepackage[normalem]{ulem}
\begin{document}

\title[DIGMN]{DIGMN: Dynamic Intent Guided Meta Network for Differentiated User Engagement Forecasting in Online Professional Social Platforms}


\author{Feifan Li}
\authornote{Work done during an internship at LinkedIn.}
\email{feifan@mail.dlut.edu.cn}
\affiliation{%
  \institution{Dalian University of Technology}
  \country{}
}

\author{Lun Du}
\authornote{Corresponding author}
\email{lun.du@microsoft.com}
\affiliation{%
  \institution{Microsoft Research}
  \country{}
}

\author{Qiang Fu}
\email{qifu@microsoft.com}
\affiliation{%
  \institution{Microsoft Research}
  \country{}
}

\author{Shi Han}
\email{shihan@microsoft.com}
\affiliation{%
  \institution{Microsoft Research}
  \country{}
}

\author{Yushu Du}
\email{yusdu@linkedin.com}
\affiliation{%
  \institution{LinkedIn Corporation}
  \country{}
}

\author{Guangming Lu}
\email{glu@linkedin.com}
\affiliation{%
  \institution{LinkedIn Corporation}
  \country{}
}

\author{Zi Li}
\email{zili@linkedin.com}
\affiliation{%
  \institution{LinkedIn Corporation}
  \country{}
}

\renewcommand{\shortauthors}{Feifan Li et al.}

\begin{abstract}
User engagement prediction plays a critical role in designing interaction strategies to grow user engagement and increase revenue in online social platforms. Through the in-depth analysis of the real-world data from the world's largest professional social platforms, i.e., LinkedIn, we find that users expose diverse engagement patterns, and a major reason for the differences in user engagement patterns is that users have different intents. That is, people have different intents when using LinkedIn, e.g., applying for jobs, building connections, or checking notifications, which shows quite different engagement patterns. Meanwhile, user intents and the corresponding engagement patterns may change over time. Although such pattern differences and dynamics are essential for user engagement prediction, differentiating user engagement patterns based on user dynamic intents for better user engagement forecasting has not received enough attention in previous works. In this paper, we proposed a \textbf{D}ynamic \textbf{I}ntent \textbf{G}uided \textbf{M}eta \textbf{N}etwork (DIGMN), which can explicitly model user intent varying with time and perform differentiated user engagement forecasting. Specifically, we derive some interpretable basic user intents as prior knowledge from data mining and introduce prior intents to explicitly model dynamic user intent. Furthermore, based on the dynamic user intent representations, we propose a meta-predictor to perform differentiated user engagement forecasting. Through a comprehensive evaluation of LinkedIn anonymous user data, our method outperforms state-of-the-art baselines significantly, i.e., 2.96\% and 3.48\% absolute error reduction, on coarse-grained and fine-grained user engagement prediction tasks, respectively, demonstrating the effectiveness of our method.
\end{abstract}




\maketitle

\section{INTRODUCTION}
Online professional social platforms like LinkedIn have become a significant part of today's lives. People use these platforms to socialize, apply for jobs, read industry news, etc. Maintaining a high-level user engagement is vital for these platforms, which can lead to more revenue (e.g., more ad exposure). For the purpose of increasing user engagement in the future, these platforms need to formulate appropriate user interaction strategies, such as delivering content that satisfies user intents or interests. Accurate user engagement prediction is one of the core technologies for developing these strategies, which can help the platform conduct user modeling, understand user needs, and provide personalized services.

\begin{figure}
    \includegraphics[width=3in, keepaspectratio]{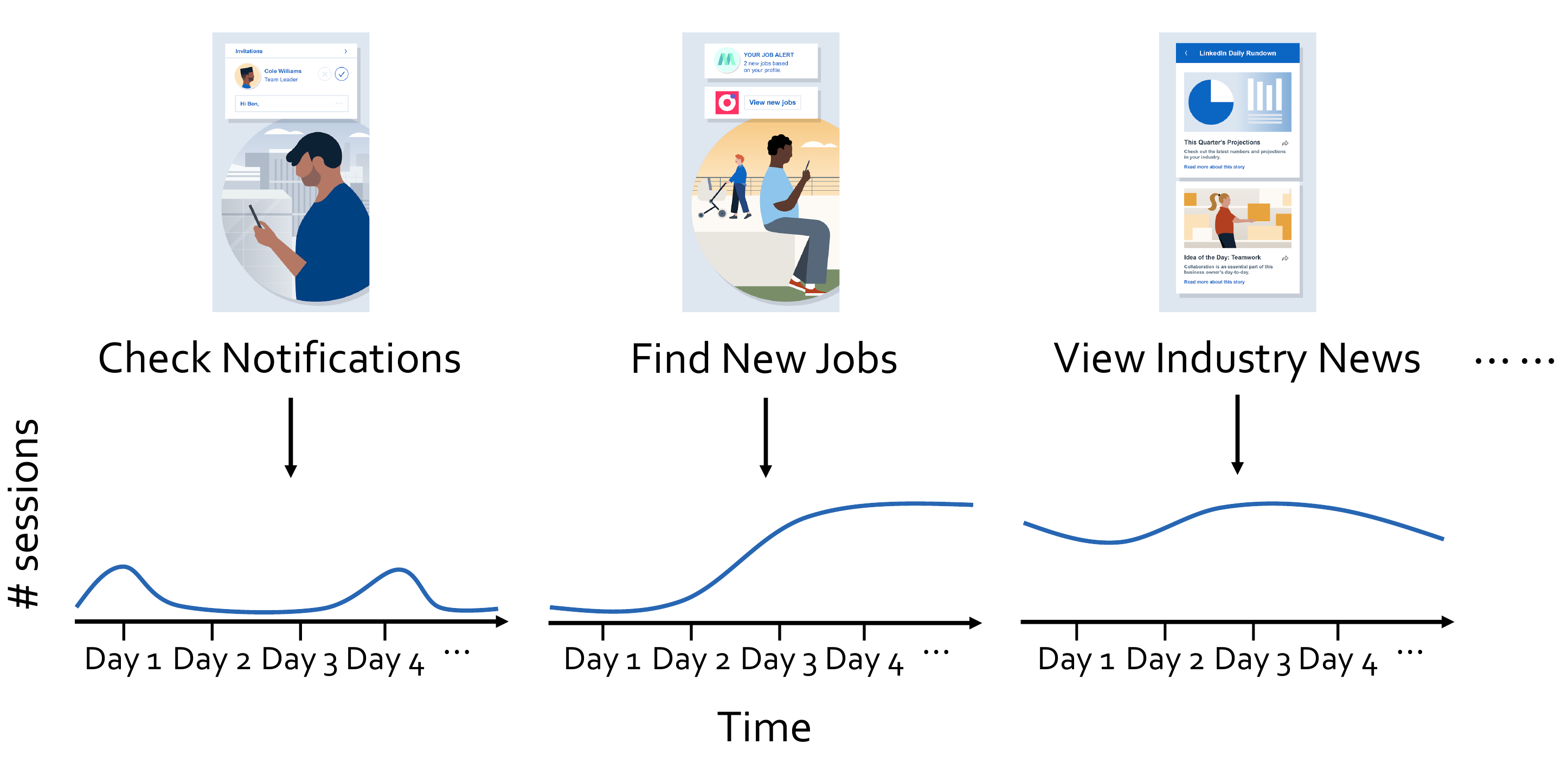}
    \caption{Example of different user engagement patterns with different user intents.}
    \label{different_engagement_pattern}
    \vspace{-0.6cm}
\end{figure}

Real-world users often exhibit different behaviors in online social platforms, leading to diverse engagement patterns. Through data mining and analysis in the scenario of LinkedIn, we found that the diversity of user engagement patterns is related to the multiple intents of users using LinkedIn, as shown in Figure \ref{different_engagement_pattern}. For instance, some users intend to look for jobs recently, and they will frequently visit LinkedIn to seek and apply for jobs, increasing their engagement level rapidly in a period. As another example, some users use LinkedIn to view industry news. Their engagement pattern is usually maintained at a relatively high level because they regularly check the industry news on LinkedIn. These insights suggest that user intents can be signals to differentiate user engagement patterns.

However, user intents are usually not directly observable because they exist implicitly in human consciousness. How to extract implicit user intents is a challenging problem. At the same time, the user's intent may change over time. For example, some users initially use LinkedIn to look for jobs, and when they finish looking for jobs, they may use LinkedIn to socialize (e.g., make new connections at a new company). Explicitly modeling dynamic user intent is vital. It can help the platform understand users' recent intents (or interests) and provide users with content that matches their intents to increase user engagement.

Recently, there have been many works on user engagement forecasting in social network platforms. \cite{yang2018know} firstly groups new users into some clusters and then uses an LSTM-based model to predict user churn rate. \cite{liu2019characterizing} constructs a user action graph to characterize and forecast new user engagement. \cite{tang2020knowing} considers user interaction actions and builds a user graph that evolves to predict user engagement. Although engagement patterns vary between users, these works use a model with static parameters for all users to predict their future engagement, which cannot sufficiently model diverse user engagement patterns and perform differentiated user engagement forecasting. Meanwhile, some works show that user intent can impact user engagement. \cite{lo2016understanding} shows that user intent can influence user engagement (e.g., usage time and return time) at Pinterest. \cite{lin2018ll} demonstrates that user primary intents are associated with how likely the user is to re-engage in activity-tracking applications. On the one hand, these works have limitations in extracting user intent. In \cite{lo2016understanding}, the user's intent is obtained through a survey when the app is just opened, which may affect the user's subsequent behaviors \cite{morwitz1993does}. \cite{lin2018ll} adopts the activity that the user most commonly uses as a proxy for the user's intent. However, the user's intent may be diverse when using applications \cite{fishbein1977belief}. On the other hand, these works do not explicitly model changes in user intent over time.

To address the above challenges, we first use Latent Dirichlet Allocation (LDA) \cite{blei2003latent} to perform user intent mining on large-scale session data and identify basic user intents. Then, we propose a  \underline{D}ynamic \underline{I}ntent \underline{G}uided \underline{M}eta \underline{N}etwork (DIGMN), which can capture the user's dynamic intent and perform differentiated user engagement prediction. Specifically, DIGMN infers multiple user intents during each session based on similarity computation with the basic user intents and captures the variation of user intents over time by the sequence model. Besides, DIGMN contains a prediction network based on meta-learning, which adopts a dynamic intent guided attention mechanism to adjust network parameters by performing a linear combination of basic parameters shared by all users for differentiated user engagement forecasting. Extensive experiments conducted on coarse-grained and fine-grained user engagement forecasting tasks verify the effectiveness of our DIGMN method.

The major contributions of this paper can be summarized as follows:
\begin{itemize}[leftmargin=*, itemsep=2pt, topsep=0pt, parsep=0pt]
\item[$\bullet$] We find that user intent can be beneficial for differentiated user engagement forecasting in online professional social platforms.
\item[$\bullet$] We develop a \textbf{D}ynamic \textbf{I}ntent \textbf{G}uided \textbf{M}eta \textbf{N}etwork (DIGMN) which explicitly model user intent's evolution over time and leverage intent guided attention mechanism to adjust model parameters for differentiated user engagement modeling and forecasting.
\item[$\bullet$] Through evaluation experiments on anonymous data from Linked\-In, our proposed model DIGMN has improvements of 2.96\% (Macro F1-score) and 3.48\% (AUROC) on coarse-grained and fine-grained user engagement prediction tasks when compared to the state-of-the-art model, showing the effectiveness of our proposed method.
\end{itemize}

\section{RELATED WORK}
In this section, we present related work on user engagement forecasting, user intent modeling, and meta-learning for dynamic network parameters.

\textbf{User engagement forecasting.} Recently, there have been many works on user engagement predicting from different perspectives on the social platform. Such as user behaviors and social attributes \cite{yang2018know}, user action graphs \cite{liu2019characterizing}, interaction actions between users \cite{tang2020knowing}, periodicity of user behaviors \cite{chowdhury2021ceam}, and causal effects of social influence \cite{zhang2022counterfactual}. These works learn a model with static parameters for all users to make predictions, which cannot sufficiently model differentiated user engagement patterns. \cite{de2018new} leverages a decision tree model to divide users into disjoint groups and then learns a separate Logistic regression model for each group of users to predict user churn. However, such separate modeling compromises the model's ability to capture similarities between users. \cite{yang2017personalized} adopts the matrix factorization to predict the personalized user's participation in mobile video. However, our scenario has a large amount of user behavior sequence data. This method can not effectively deal with the user's behavior sequence and its change over time.

\textbf{User intent modeling.} Some previous works exploit LDA \cite{xu2008modelling, carman2010towards, liu2019characterizing}, n-gram \cite{lin2009modeling, liu2019characterizing} and deep learning model \cite{loyola2017modeling, agrawal2018learning} to mine user intent from user behavior. At the same time, modeling user intents can help us understand user needs better and is significant in many scenarios: for instance, web searching  \cite{guo2010ready, gu2022accelerating}, e-commerce application  \cite{lo2016understanding}, image sharing social platform  \cite{cheng2017predicting}, activity tracking application \cite{lin2018ll} and recommender systems  \cite{li2021intention, chen2022intent, kota2021learnings, wang2019tag2gauss, wang2019tag2vec}. However, to our best knowledge, no related work has explicitly modeled user intent and its variation for differentiated user engagement forecasting in online social platforms.

\textbf{Meta-learning for dynamic network parameters.} Meta-lear\-ning (also known as learning to learn) can be used to learn dynamic model parameters, which is widely used in scenarios and tasks with diverse data distributions. There are usually two ways to learn dynamic model parameters \cite{han2021dynamic}: dynamically generate model parameters or dynamically adjust model parameters.
For dynamically generating model parameters: \cite{pan2019urban} utilizes a meta-knowledge learner to generate model parameters for modeling diverse spatial-temporal correlations in urban traffic prediction. \cite{zhu2022personalized} adopts a meta-network to learn a personalized mapping function for each user in the cross-domain recommendation. \cite{zhang2022leaving}, \cite{yan2022apg} and \cite{bian2022can} adopt meta networks to generate dynamic parameters of scenario-specific models for CTR prediction. 
For dynamically adjusting model parameters: \cite{zhang2017supplementary} incorporates meta-information learned from a supplementary neural network into a fixed base-level neural network to realize the generalization for each type of input image. \cite{yang2019condconv}, \cite{chen2020dynamic}, and \cite{li2022omni}, in the light of different input images, use a separate network to learn different weights for adjusting the parameters of convolution kernels, which can obtain dynamic convolution kernels for feature extraction. However, no existing works utilize meta-learning for predicting user engagement with diverse patterns in online social platforms.

\section{Preliminary}
In this section, we introduce the relevant definitions and formulate the user engagement forecasting tasks used in this work.
\subsection{Basic Definitions}
We define $U=\{u_{1}, u_{2},..., u_{N}\}$ as a set of different users and $I=\{i_{1}, i_{2}, ..., i_{M}\}$ as a set of different event types, where $N, M$ are the number of users and event types respectively.

\textbf{User event.} The user's behaviors on the platform can be abstracted as events for collection. A user event $e$ can be denoted by $e=(u, i, t)$, where $u\in U$ is the user, $i\in I$ is the event type, and $t$ is the time when the event happens. We collect ten major event types on the platform, as shown in Table \ref{Types of user events we collect on LinkedIn}. These ten events can completely cover users' activities on the platform.

\begin{table}[H]
    \setlength{\abovecaptionskip}{0.2cm}
    \setlength{\belowcaptionskip}{0.2cm}
    \vspace{-0.4cm} 
    \small
    \centering
    \caption{Types of user events we collect on the platform.}
    \label{Types of user events we collect on LinkedIn}
    \begin{tabular}{|c|c|c|}
        \hline Event ID & Event Type & Explanation\\
        \hline 1 & Feed & view and react to updates\\
        \hline 2 & Search & search for members, jobs, or other\\
        \hline 3 & View Profile & view profiles of members or companies\\
        \hline 4 & Jobs & view or apply for jobs\\
        \hline 5 & PYMK & invite members to build connections\\
        \hline 6 & Notification & check notifications\\
        \hline 7 & Message & check or send messages\\
        \hline 8 & Edit Profile & edit personal profile\\
        \hline 9 & Share Content & share content\\
        \hline 10 & Follow & follow members, companies, or other\\
        \hline
    \end{tabular}
    \vspace{-0.2cm} 
\end{table}

\textbf{User session.} A user session can be viewed as a set of continuous browsing activities with gaps between full-page views not more than a threshold. Formally, a session can be denoted as $\mathcal{S}=(u, \mathbf{c}, \mathcal{E})$, where $u\in U$ is the user who starts the session,  $\mathbf{c}$ is the session context information (e.g., the session start time, the session duration time and the software client) and $\mathcal{E}$ is the sequence of events that take place during the session.

\textbf{User engagement.} User engagement measures how often and how long users interact with the website or application, reflecting how much value users get from the website or application.
Different platforms use different metrics to track user engagement. In this paper, we adopt two metrics to measure the engagement of each user on LinkedIn: average active days (coarse-grained, which is the same as the definition of active rate in \cite{liu2019characterizing}) and average session numbers (fine-grained). The average session numbers $\bar{s}=\frac{\#\ valid\ sessions}{\#\ total\ days}$, where a valid session is a session that contains at least one event in Table \ref{Types of user events we collect on LinkedIn}. The average active days $\bar{d}=\frac{\#\ active\ days}{\#\ total\ days}$, where an active day is defined as a day that the user has at least one valid session.

\textbf{Platform actions.} Previous studies have demonstrated that user engagement on the platform can be increased through notifications \cite{alkhaldi2016effectiveness, yuan2019state}, incentives including badges \cite{anderson2013steering, anderson2014engaging}, etc. Therefore, we need to consider the impact of the interaction actions between the platform and the user when forecasting user engagement. 
For simplicity, we only consider the platform delivery messages. 
Formally, a platform delivery message can be denoted as $\mathcal{D}=(u, w, r, t)$, where $u\in U$ is the user being delivered, $w$ is the way the message is delivered (i.e., email, SMS and in-app push), $r$ is the content of the delivery message (i.e., Feed relevant, Jobs relevant, PYMK relevant, Message relevant and others), $t$ is the delivery time.

\subsection{Task Formulation}
In this work, we predict the trend of user engagement instead of directly predicting the future engagement of users because, in actual business scenarios, we pay more attention to the change in user engagement. For example, when designing a platform message delivery strategy in the scenario of user retention, the change value ($\Delta\overline{s}$ or $\Delta\overline{d}$) of the user engagement caused by the delivered message is an essential indicator of business decisions. Specifically, we define two user engagement trend forecasting tasks as follows:
\begin{itemize}[leftmargin=*, itemsep=2pt, topsep=0pt, parsep=0pt]
\item[$\bullet$] \textbf{Day-level Task (coarse-grained)}: We predict the trend of the average active days of the user, which can be viewed as a 3-class (i.e., increase, decrease, or stay the same) classification task.
\item[$\bullet$] \textbf{Session-level Task (fine-grained)}: We predict the trend of the average session numbers of the user, which can be viewed as a 2-class (i.e., decrease or not decrease) classification task. The reason for setting it as a binary classification task is that the number of user sessions has a broader distribution than the number of active days, and users with a large number of sessions are less likely to have the same number of sessions over a while.
\end{itemize}

Based on the above definitions, for any user $u\in U$, given the user's macroscopic features $\bm{M}_{u}$ (e.g., the number of connections, the average number of sessions per day in the past period), session sequence $\bm{S}_{u}=<\mathcal{S}_{1}, \mathcal{S}_{2}, ..., \mathcal{S}_{T}>$, and the latest platform delivery message $\bm{D}_{u}$ in the past period, forecasting user engagement trend $y_{u}$ in the next period can be formulated as follows:
\begin{equation}
    {\rm Pr}(y_{u}|\bm{M}_{u}, \bm{S}_{u}, \bm{D}_{u},; \Theta)
\end{equation}
where $T$ is the number of sessions, $y_{u}$ is the user engagement trend label, and $\Theta$ is the parameters we need to learn.

\section{Intent mining}
In this section, we introduce how to perform user intent mining and show the basic user intents we obtain from data mining.

We assume that every user has at least one intent within each session \cite{ajzen1991theory}. The user's intents are usually not directly observed and exist implicitly in the user's consciousness. It is difficult for us to collect a large number of user behavior samples with user intent labels, so using unsupervised methods for user intent mining is a more feasible option.
Intuitively, the user's intents influence the user's behaviors, and the user's behaviors constitute the set of events in the session. This process is similar to generating documents with an unsupervised topic model LDA. Like previous work \cite{liu2019characterizing}, we treat each session as a document and each event type as a word, then apply LDA to mine basic user intents.

We use Spark MLlib\footnote{https://spark.apache.org/mllib/} to perform LDA on approximately 6 million anonymous user session data for intent mining. To determine the optimal number of intents with semantic meaning, we adopt perplexity as the evaluation metric to search for the optimal intent number in $\{2, 3, 4, ..., 10\}$. When the number of intents equals 7, the LDA model has the least perplexity.

\begin{figure}[H]
    \setlength{\abovecaptionskip}{0.cm}
    \vspace{-0.4cm}
    \centering
    \includegraphics[width=3in, keepaspectratio]{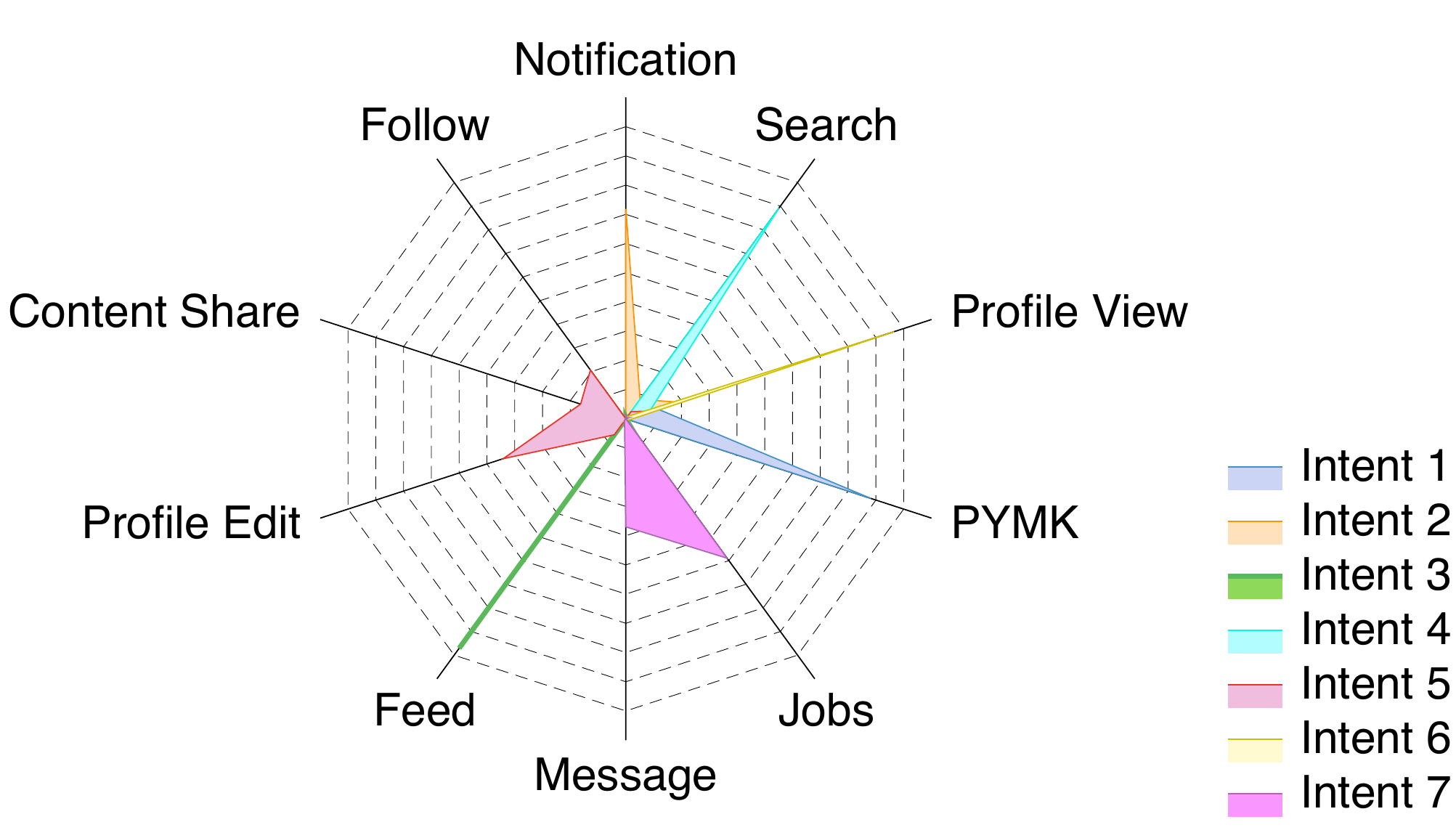}
    \caption{Basic user intents. LDA obtains 7 basic intents, and we show the event weights that compose them. Best viewed in color.}
    \label{7 Intent}
    \vspace{-0.4cm}
\end{figure}

These 7 topics can be regarded as 7 basic user intents (denoted as $\bm{t}_{1}, \bm{t}_{2}, ..., \bm{t}_{7}\in\mathbb{R}^{10}$), each of which is composed of events with different weights (also can be view as probability) as shown in Figure \ref{7 Intent}. The meaning of each intent can be explained by the top-weight events that make up them. For example, we can find that the two events with top weight making up intent 1 are PYMK and Profile View, indicating that the users want to expand their connections on LinkedIn.

\section{Dynamic Intent Guided Meta Network}
\begin{figure*}
    \vspace{-0.4cm}
    \centering
    \includegraphics[width=5.0in, keepaspectratio]{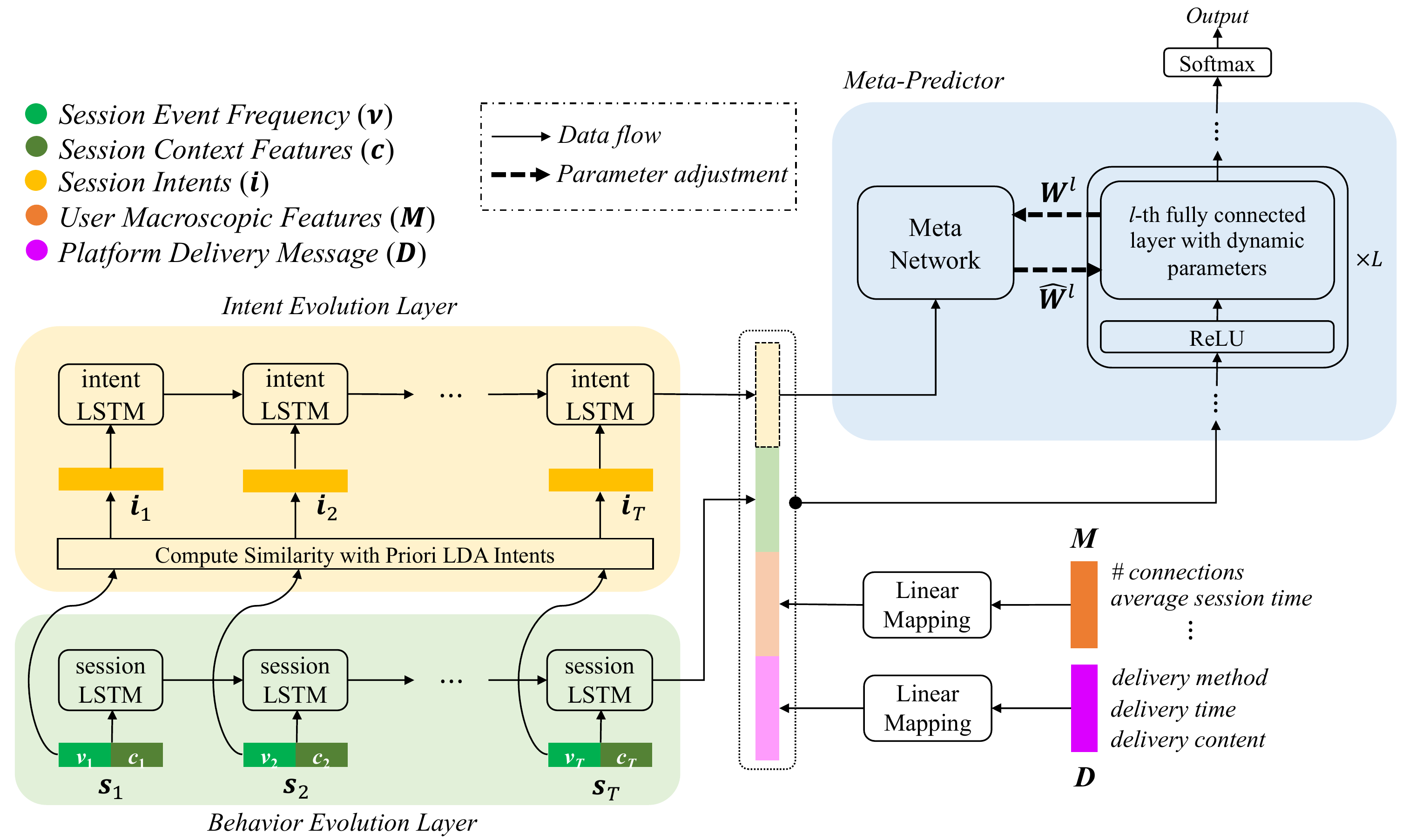}
    \caption{The framework of our proposed model. It consists of three main components: a behavior evolution layer, an intent evolution layer, and a meta-predictor. Best viewed in color.}
    \label{Model}
    \vspace{-0.4cm}
\end{figure*}

In this section, we introduce our proposed model, DIGMN. The framework of our proposed model is illustrated in Figure \ref{Model}. It consists of three main components: a behavior evolution layer, an intent evolution layer, and a meta-predictor.

\subsection{Behavior Evolution Layer}

Users' behaviors change over time, and the length of the behavior sequence of different users may be different. LSTM\cite{hochreiter1997long} has been widely used for modeling variable-length user behavioral sequences. In this paper, we utilize a 1-layer LSTM to model user behavior evolution.

For each session $\mathcal{S}$, we represent it by concatenating the session context features $\bm{c}$ and session event frequency $\bm{\nu}=(\nu_{1}, \nu_{2}, ..., \nu_{10})\in\mathbb{R}^{10}$, where  $\nu_{i}$ is the frequency of event $i$ (e.g., if there are 100 events in a session, and the Feed event occurs 10 times, then $\nu_{1}=10/100=0.1$). We take $\bm{s}=(\bm{\nu},\bm{c})$ as the input of the session LSTM unit at each timestep. Since the user's behavior sequences are of different lengths, we take the last timestep output of LSTM as the representation of the user's behavior features in the past period.

\subsection{Intent Evolution Layer}
A user may have multiple intents during one session. For example, the users can use LinkedIn to view updates on industry news (i.e., intent 3) and share content related to themself (i.e., intent 5) at the same time. We can infer the user's possible intents based on the events happening during this session.

In this paper, we exploit a simple and effective way to infer user intent in each session. Given a session's event frequency $\bm{\nu}$ and the basic user intents $\bm{t}_{1}, \bm{t}_{2}, ..., \bm{t}_{7}$ obtained by LDA in the intent mining stage, we can infer the user intents $\bm{i}=(i_{1}, i_{2}, ..., i_{7})$ during this session. To be specific, $i_{k}$ is the cosine similarity between the session event frequency $\bm{\nu}$ and the $k$-th basic user intent $\bm{t}_{k}$, which measures the degree to which the current session contains the $k$-th intent and can be calculated as follows:
\begin{equation}
    i_{k}=\frac{\bm{\nu}\cdot \bm{t}_{k}}{||\bm{\nu}||\times ||\bm{t}_{k}||}
\end{equation}

By performing the above intent extraction operation on each user session, we can obtain an intent sequence for each user with the same length as the user session sequence. Another 1-layer LSTM is adopted to model the variation of user intent over time. We take each session's intent representation $\bm{i}$ as the input of the intent LSTM unit at each timestep. We also take the LSTM's last timestep output to represent the user's dynamic intent over the past period.

\subsection{Meta-Predictor}
Different users have different engagement patterns, and the user engagement pattern may change in different periods. Previous works \cite{yang2018know, liu2019characterizing, tang2020knowing} use a model with the static parameters $\Theta$ for different users to predict their future engagement in the social platforms, which can be formulated as $\bm{y}=F(\bm{x}, \Theta)$, where $\bm{x}$ is the model input (i.e., user features) and $\bm{y}$ is the model output. To make a good prediction, the model tends to capture the common patterns of all users, which is not optimal for modeling and predicting diverse user engagement patterns in real scenarios. The most straightforward idea is to learn a unique predictor for each user or a group of similar users. However, there is a certain similarity between users, while separate modeling may impair the model's ability to capture the similarity between users. Therefore, we need a predictor that can better model the diversity in user engagement patterns while capturing such similarity.

Dynamic parameters neural networks have shown promising results in various real-world tasks and can dynamically adjust or generate model parameters $\Theta^{*}$ according to different inputs $\bm{x}$, which have a more powerful representation ability. It can be formulated as $\bm{y} = F(\bm{x}, \Theta^{*})=F(\bm{x}, \phi(\bm{x}, \Theta))$, 
where $\phi(\cdot, \Theta)$ is the operation that adjusts or generates model parameters according to input $\bm{x}$. A common technique in dynamic parameters neural networks is attention on parameters \cite{yang2019condconv, chen2020dynamic}. It assumes that there are some basic learnable parameters with the same shape in the dynamic parameter layer. Given different inputs, the meta network can generate different attention weights to combine these basic learnable parameters in the dynamic parameter layer to obtain dynamic parameters and make transformations. In our scenario, the meta network can adjust the model parameters according to different user feature inputs for differentiated modeling. At the same time, all users share the basic learnable parameters of the dynamic parameter layer, which enables the model to capture the similarities between users.
 
According to the previous analysis, we can use the user dynamic intent representation obtained from the intent evolution layer as the input signal to parameterized adjustment operation $\phi(\cdot, \Theta)$ in the meta-network to perform differentiated user engagement forecasting. Specifically, we design a meta-predictor that contains a meta network and multiple stacked fully connected layers with dynamic parameters (FC-D layer), as illustrated in Figure \ref{Model}.

\subsubsection{\textbf{Fully Connected Layers with Dynamic Parameters}}
\begin{figure}
    \vspace{-0.4cm}
    \setlength{\abovecaptionskip}{0.cm}
    \centering
    \includegraphics[width=2.6in, keepaspectratio]{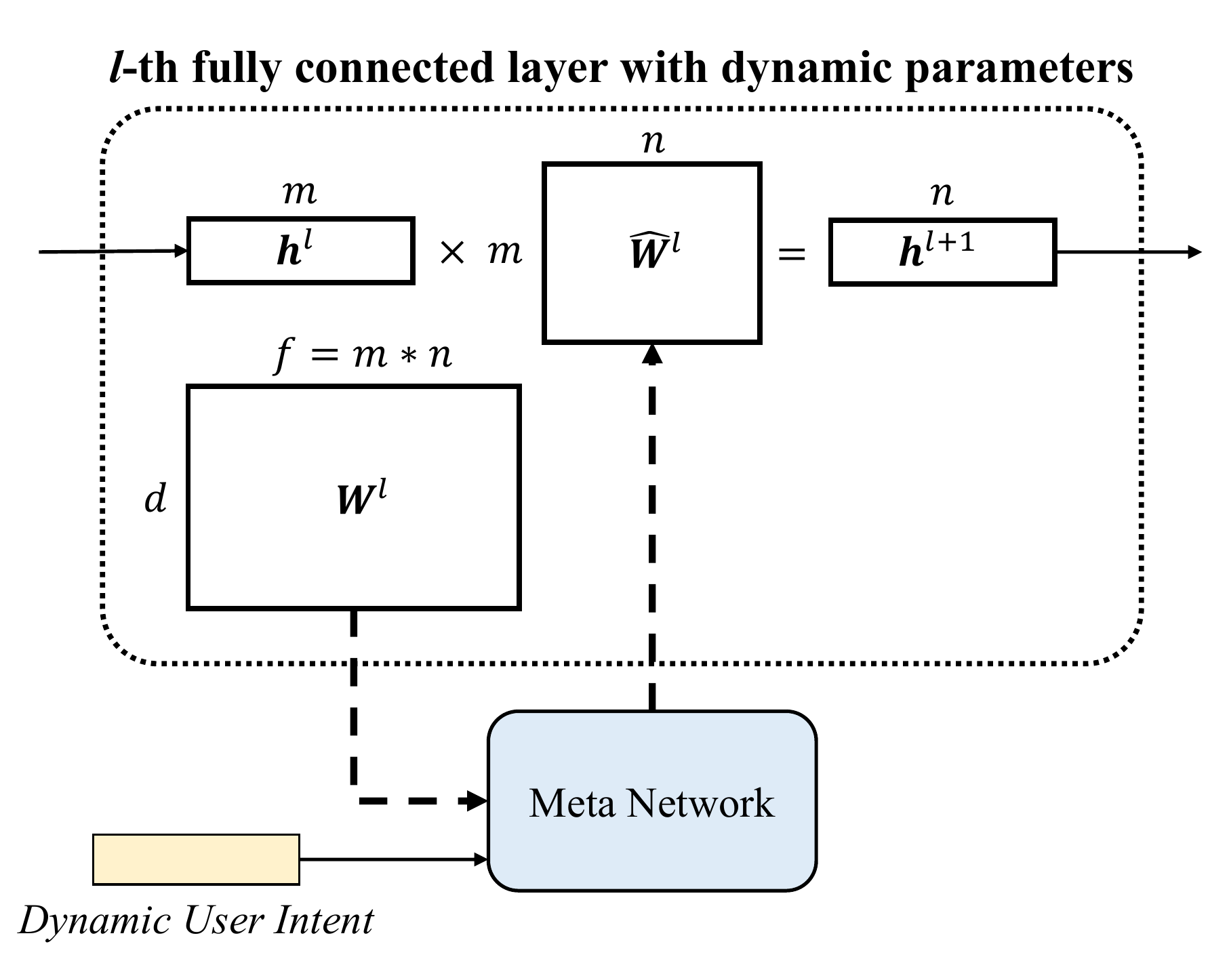}
    \caption{The $l$-th fully connected layers with dynamic parameters.}
    \label{DFC}
    \vspace{-0.4cm}
\end{figure}

As illustrated in Figure \ref{DFC}, assuming that the $l$-th FC-D layer transforms the input feature $\bm{h}^{l}\in\mathbb{R}^{m}$ into $\bm{h}^{l+1}\in\mathbb{R}^{n}$, and correspondingly assuming there are $d$ basic learnable parameters $\bm{W}^{l}_{1}, \bm{W}^{l}_{2}, ..., \bm{W}^{l}_{d}\in\mathbb{R}^{(m*n)}$ in it, denoted as $\bm{W}^{l}=[\bm{W}^{l}_{1}, \bm{W}^{l}_{2}, ..., \bm{W}^{l}_{d}]^{T}\in\mathbb{R}^{d\times(m*n)}$. The $l$-th FC-D layer computes the following transformation:
\begin{equation}
    \begin{aligned}
    \bm{h}^{l+1}=\bm{\widehat{W}}^{l}\cdot\bm{h}^{l}+\bm{\hat{b}}^{l}
    \end{aligned}
\end{equation}
where $\bm{\widehat{W}}^{l}\in\mathbb{R}^{m\times n}, \bm{\hat{b}}^{l}\in\mathbb{R}^{n}$ are the weights and bias of the $l$-th FC-D layer adjusted by the meta network.

\subsubsection{\textbf{Meta Network}}
\begin{figure}
    \vspace{-0.4cm}
    \setlength{\abovecaptionskip}{0.cm}
    \centering
    \includegraphics[width=2.5in, keepaspectratio]{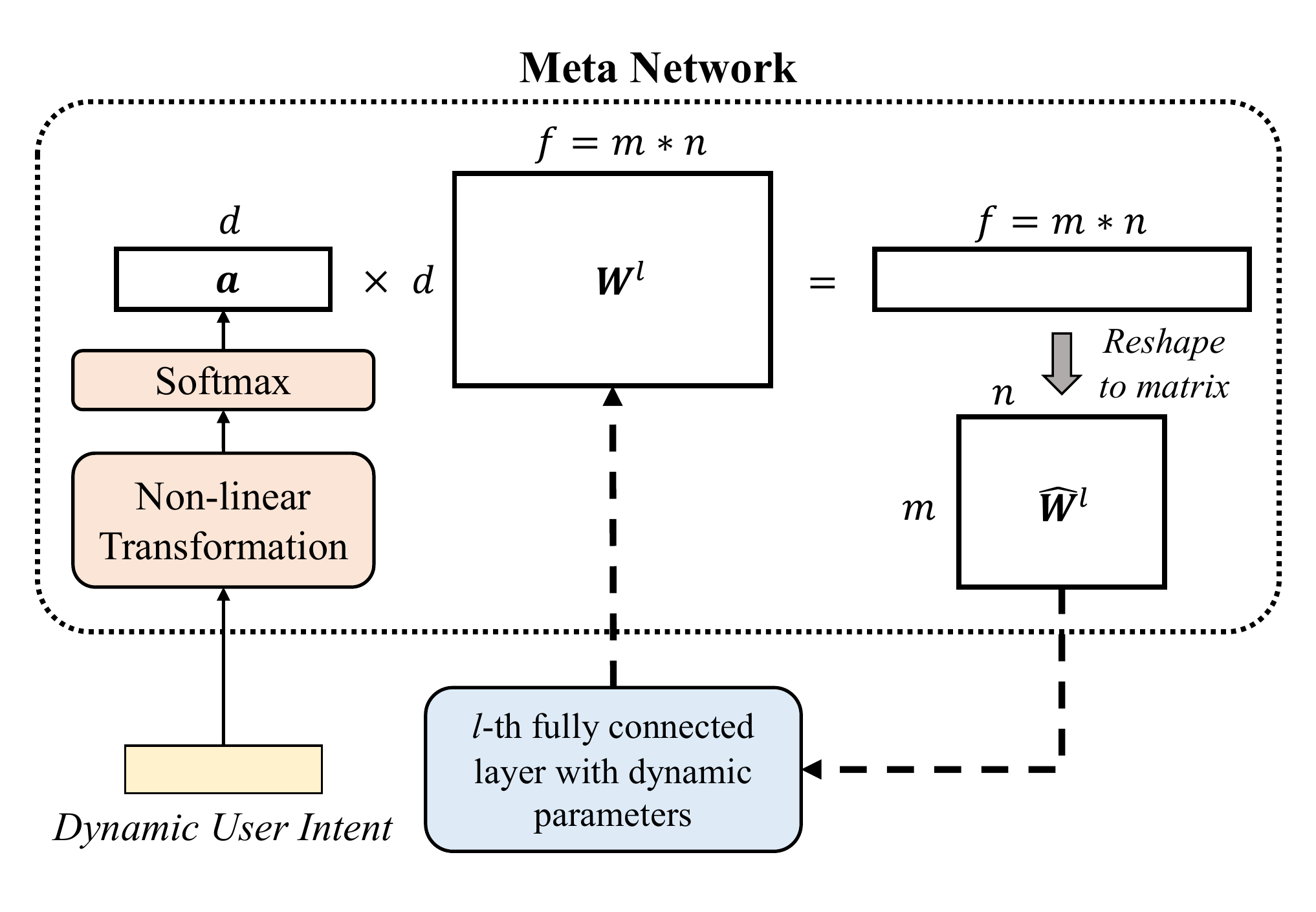}
    \caption{Meta Network.}
    \label{PAN}
    \vspace{-0.4cm}
\end{figure}

The adjustment process of $\bm{\widehat{W}}^{l}$ in the meta network is shown in Figure \ref{PAN}. The combination weights $\bm{a}=[a_{1},..., \\ a_{d}]^{T}\in\mathbb{R}^{d}$ (which can be seen as meta-knowledge) of basic learnable parameters $\bm{W}^{l}$ is derived by applying a non-linear transformation and a Softmax operation to user dynamic intent (denoted as $\bm{\widetilde{i}}$). In this paper, we utilize two fully connected layers with the ReLU activation function to perform non-transformation, which can be formulated as follows:
\begin{equation}
    \begin{aligned}
        \bm{a}={\rm Softmax}(\bm{W}_{2}({\rm ReLU}(\bm{W}_{1}\bm{\widetilde{i}}+\bm{b}_{1}))+\bm{b}_{2})
    \end{aligned}
\end{equation}
where $\bm{W}_{1}, \bm{W}_{2}, \bm{b}_{1}, \bm{b}_{2}$ are learnable parameters. Then we can combine the basic learnable parameters $\bm{W}^{l}=[\bm{W}^{l}_{1}, \bm{W}^{l}_{2}, ..., \bm{W}^{l}_{d}]^{T}$ through $\bm{a}$, and obtain the adjusted transformation matrix $\bm{\widehat{W}}^{l}$ of $l$-th FC-D layer through the reshape operation:
\begin{equation}
    \bm{\widehat{W}}^{l}={\rm Reshape}(\bm{a}^{T}\bm{W}^{l})={\rm Reshape}(\sum\nolimits_{i=1}^{d}a_{i}\bm{W}_{i}^{l})
\end{equation}
$\bm{\hat{b}}^{l}$ can be calculated in the similar way as $\bm{\widehat{W}}^{l}$.

Each row of $\bm{W}^{l}$ (i.e., $\bm{W}^{l}_{i}, i=1,2,..,d$) can be regarded as independent basic learnable parameters. $\bm{\widehat{W}}^{l}$ is a linear combination of $\bm{W}^{l}_{1}, \bm{W}^{l}_{2}, ..., \bm{W}^{l}_{d}$, and we hope that $\bm{W}^{l}_{1}, \bm{W}^{l}_{2}, ..., \bm{W}^{l}_{d}$ are as orthogonal as possible to reduce redundant information. \cite{bansal2018can} summarizes some orthogonal regularization methods for parameters, including "selective" Soft Orthogonality Regularization, which can be written as follows:
\begin{equation}
    \lambda\cdot||\bm{W}^{l}(\bm{W}^{l})^{T}-\bm{I}||^{2}_{F}
\end{equation}
where $\lambda$ is the regularization coefficient, $\bm{I}\in\mathbb{R}^{d\times d}$ is the identity matrix. Here we do not limit the norm of $\bm{W}^{l}_{i}$ to 1, and adopt the following orthogonal regularization term:
\begin{equation}
    \mathcal{L}_R=\sum\nolimits_{l=1}^{L}||(\bm{W}^{l}(\bm{W}^{l})^{T}-\bm{I})\odot(\textbf{1}-\bm{I})||^{2}_{F}
\end{equation}
where $\bm{1}\in\mathbb{R}^{d\times d}$ is a matrix whose elements are all 1, $\odot$ is Hadamard product, and $L$ is the number of FC-D layers.

\subsection{End-to-end model training}
We adopt an end-to-end learning strategy and use error backpropagation to train our proposed model. We choose cross-entropy as a classification loss function, which can be formulated as follows:
\begin{equation}
    \mathcal{L}_C=-\frac{1}{n}\sum\nolimits_{i}\sum\nolimits_{c}y_{ic}\log(\hat{y}_{ic})
\end{equation}
where $n$ is the number of samples, $i$ represents the $i$-th sample, $y_{ic}$ equals to 1 if $i$ belongs to class $c$ otherwise 0, $\hat{y}_{ic}$ is the predicted probability that sample $i$ belongs to class $c$.  
After adding the regularization term for the parameters of the meta-predictor, the loss function can be written as:
\begin{equation}
    \mathcal{L}=\mathcal{L}_{C}+\beta\cdot\mathcal{L}_{R}
\end{equation}
where $\beta$ is the hyperparameter that trades off classification loss $\mathcal{L}_{C}$ and regularization term $\mathcal{L}_{R}$.

\section{Evaluation}
In this section, we describe the dataset used in this work and introduce the detailed experimental settings. Then, we show that DIGMN outperforms current state-of-the-art models on the user engagement prediction task, demonstrating the effectiveness of DIGMN. To be more specific, we aim to answer the following research questions:
\begin{itemize}[leftmargin=*, itemsep=2pt, topsep=0pt, parsep=0pt]
\item[$\bullet$] \textbf{RQ1}: Can DIGMN outperform state-of-the-art baselines in user engagement forecasting tasks at different granularities?
\item[$\bullet$] \textbf{RQ2}: Does DIGMN perform better than a network with static parameters and a similar number of parameters?
\item[$\bullet$] \textbf{RQ3}: How does each part in DIGMN affect the performance?
\item[$\bullet$] \textbf{RQ4}: Is dynamic user intent a good signal to differentiate diverse user engagement patterns?
\item[$\bullet$] \textbf{RQ5}: How do hyperparameters affect the performance?
\item[$\bullet$] \textbf{RQ6}: What about the impact of using other methods to obtain user intent representations on DIGMN performance?
\item[$\bullet$] \textbf{RQ7}: What is the influence of different methods implementing dynamic parameters on the prediction effect of DIGMN?
\item[$\bullet$] \textbf{RQ8}: Can DIGMN learn interpretable user dynamic intent representations?
\end{itemize}

\subsection{\textbf{Experimental Setup}}
\subsubsection{\textbf{Datasets}} To evaluate the performance of our proposed model, we conduct experiments on real-world anonymous users' data from LinkedIn. To protect user privacy, we collect coarse-grained behavioral data (only the types of events, as shown in Table \ref{Types of user events we collect on LinkedIn}, not detailed user behaviors) of random anonymous users during four weeks. We filter out users with less than 7 (median) sessions during the first two weeks to make the extracted dynamic intent information more meaningful. As shown in Table \ref{Label}, the user engagement trend label $y$ is obtained by comparing the user engagement in the first two weeks (i.e., $\overline{d}_{h}$, $\overline{s}_{h}$) and the following two weeks (i.e., $\overline{d}_{f}$, $\overline{s}_{f}$).
The label distribution of the day-level task is approximately $y=-1:y=0:y=1\approx2:2:1$, and the label distribution of the session-level task is approximately $y=0:y=1\approx3:2$.
We randomly sample 200K users for each experiment and split them into three parts: 80\% of the samples (i.e., users) for training, 10\% for validation, and the rest 10\% for testing.

\begin{table}[H]
  \setlength{\abovecaptionskip}{0.2cm}
  \setlength{\belowcaptionskip}{0.2cm}
  \vspace{-0.4cm}
  \small
  \caption{Labels of user engagement forecasting tasks.}
  \label{Label}
  \begin{tabular}{c|c}
    \toprule
    \textbf{Day-level Task} & \textbf{Session-level Task} \\
    \midrule
    $
        \textit{y}=\left\{
        \begin{aligned}
            -1, \quad &\overline{d}_{h} > \overline{d}_{f}\\
            0, \quad &\overline{d}_{h} = \overline{d}_{f}\\
            1, \quad &\overline{d}_{h} < \overline{d}_{f}
        \end{aligned}
        \right. \nonumber
    $
    &
    $
        \textit{y}=\left\{
        \begin{aligned}
            0, \quad &\overline{s}_{h} > \overline{s}_{f}\\
            1, \quad &\overline{s}_{h} \leq \overline{s}_{f}
        \end{aligned}
        \right. \nonumber
    $
    \\
    \bottomrule
  \end{tabular}
  \vspace{-0.3cm} 
\end{table}

\subsubsection{\textbf{Evaluation Metrics.}} For the day-level task (3-class classification task), we adopt Macro F1-score as our metric following Liu et al.\cite{liu2019characterizing}'s work. Macro F1-score is the average F1-score for each class and can evaluate the classifier in the case of class imbalance. Furthermore, for the session-level task (2-class classification task), AUROC is adopted as our metric, which is a popular metric for the binary classification task.

\subsubsection{\textbf{Comparison Methods.}} We do not collect the interaction actions between users in this work. Considering comparability, we did not compare with the FATE \cite{tang2020knowing} and CFChurn \cite{zhang2022counterfactual}, which uses interaction behaviors between users to predict user engagement or user churn. We compare our proposed model against the following baselines:
\begin{itemize}[leftmargin=*, itemsep=2pt, topsep=0pt, parsep=0pt]
\item[$\bullet$] \textbf{Logistic Regression (LR)}: We use user macroscopic features and the latest platform delivery message to make predictions.
\item[$\bullet$] \textbf{XGBoost} \cite{chen2016xgboost}: Use the same input features as Logistic Regression to classify.
\item[$\bullet$] \textbf{Multilayer Perceptron (MLP)}: We implement an MLP with two hidden layers 
using the same features as Logistic Regression and XGBoost as input.
\item[$\bullet$] \textbf{Activity LSTM} \cite{yang2018know}: We count the number of events in Table \ref{Types of user events we collect on LinkedIn} that occurred each day for the user in the past two weeks and get a time series $\bm{A}\in\mathbb{R}^{10\times14}$ as the input of the activity LSTM. 
\item[$\bullet$] \textbf{Temporal GCN-LSTM} \cite{liu2019characterizing}: We treat events as actions and construct the user's action graph according to \cite{liu2019characterizing}. Temporal GCN-LSTM applies GCN on the action graph to extract user fine-grained features, then feeds it into LSTM to capture its temporal dynamics. 
\item[$\bullet$] \textbf{Deep Multi-channel} \cite{liu2019characterizing}: To achieve the best performance, it combines user macroscopic features, the latest platform delivery message, activity LSTM, and Temporal GCN-LSTM. It can be viewed as the state-of-the-art model for predicting user engagement that does not consider the interaction behaviors between users. 
\end{itemize}

\subsubsection{\textbf{Evaluation Settings.}} The output dimension of the linear mapping layer is set to be half of the input dimension. The number of hidden units in session LSTM and intent LSTM is empirically set to 32. We implement DIGMN with $L$ FC-D layers, where $L$ is searched from 1 to 5. 
The hidden units of the non-linear transformation layer in the meta network are set to 32. The dimension of $\bm{a}$ is searched from 2 to 10 and the hyperparameter $\beta$ is searched in $[10, 1, 10^{-1}, 10^{-2}, 10^{-3}, 10^{-4}, 10^{-5}, 0]$. We adopt Adam \cite{kingma2014adam} as the optimizer with an initial learning rate $10^{-3}$ and learning rate decay half every 20 epochs. According to \cite{xie2017all}, the authors argue against using $\ell_{2}$ weight decay and the orthogonal regularization term together, so we only use $\ell_{2}$ weight decay for learnable parameters other than $\bm{W}^{l} (l=1,2,...,L)$ in FC-D layers with a decay rate $10^{-5}$. An early stopping strategy is also used on the validation set to avoid overfitting. Our experiments are conducted on a single machine with a 2.4GHz 12-core CPU and 64GB of memory. Each experiment is repeated 5 times with different random seeds.

\subsection{Performance Comparison (RQ1, RQ2)}
Table \ref{Performance} lists the experiment results of all compared methods. We observe that DIGMN performs best on both day-level and session-level user engagement prediction tasks, which shows the effectiveness of DIGMN. The possible reason for the poor performance of activity LSTM and Temporal GCN-LSTM on our dataset is that our user session data is sparse (each user averages around 2 sessions a day in our dataset, but 7 in Liu et al.\cite{liu2019characterizing}'s dataset), which results in sparse activity sequences and action graphs, reducing their performance.

Compared with the FC layer, the FC-D layer used in DIGMN increases the learnable parameters. Take one FC-D layer as an example, assuming that it maps the input $\bm{h}\in\mathbb{R}^{m}$ to $\bm{h'}\in\mathbb{R}^{n}$, then the learnable parameters of FC-D layer is $d\ast m\ast n$ (for brevity, ignoring bias). However, for the FC layer, its number of learnable parameters is $m\ast n$. 
For comparability, we use the FC layers (Static) with the same shape or a similar number of parameters to replace the FC-D layers (Dynamic) in DIGMN. As shown in Table \ref{DIGMN_vs_MLP}, we observe that DIGMN with dynamic parameters outperforms DIGMN with static parameters in two tasks, demonstrating the effectiveness of leveraging dynamic parameters network to achieve differentiated user engagement prediction.

\begin{table}
  \vspace{-0.2cm} 
  \small
  \caption{Performance comparison on the classification of user engagement trends at the day-level and the session-level.}
  \label{Performance}
  \begin{tabular}{lcc}
    \toprule
    \multirow{2}{*}{\diagbox{\textbf{Model}}{\textbf{Task}}} & \textbf{Day-level} & \textbf{Session-level} \\
    \cmidrule(r){2-2} \cmidrule(r){3-3} & \textbf{Macro F1-Score} & \textbf{AUROC}\\
    \midrule
    \textit{Feature-based model} \\
    \midrule
    LR & $0.446\pm0.000$ & $0.561\pm0.000$ \\
    XGBoost & $0.460\pm0.001$ & $0.569\pm0.001$\\
    \midrule
    \textit{End-to-end neural network model} \\
    \midrule
    MLP & $0.463\pm0.002$ & $0.571\pm0.002$\\
    Activity LSTM & $0.518\pm0.001$ & $0.605\pm0.001$\\
    Temporal GCN-LSTM & $0.522\pm0.002$ & $0.600\pm0.003$\\
    Deep Multi-channel & $0.575\pm0.002$ & $0.633\pm0.002$\\
    \textbf{DIGMN} (ours) & $\bm{0.592\pm0.001}$ & $\bm{0.655\pm0.001}$\\
    \midrule
    Improvements & $2.96\%(\uparrow)$ & $3.48\%(\uparrow)$ \\
    \bottomrule
  \end{tabular}
  \vspace{-0.4cm} 
\end{table}

\begin{table*}
  \vspace{-0.2cm}  
  \small
  \caption{Performance comparison of DIGMN with dynamic predictor and DIGMN with the static predictor.}
  \label{DIGMN_vs_MLP}
  \begin{tabular}{lccccc}
    \toprule
    \multirow{2}{*}{\diagbox{\textbf{Model}}{\textbf{Task}}} & \multirow{2}{*}{\textbf{\# Hidden Units}} & \multicolumn{2}{c}{\textbf{Day-level}} & \multicolumn{2}{c}{\textbf{Session-level}} \\
    \cmidrule(lr){3-4} \cmidrule(lr){5-6} & \textbf{in Predictor} & \textbf{Macro F1-Score} & \textbf{\# Parameters} & \textbf{AUROC} & \textbf{\# Parameters}\\
    \midrule
    DIGMN (Static) & \{64, 32\} & $0.580\pm0.001$ & 18.8K & $0.642\pm0.001$ & 18.7K\\
    DIGMN (Static) & \{160, 96\} & $0.585\pm0.001$ & 38.6K & $0.645\pm0.002$ & 38.5K\\
    DIGMN (Dynamic) & \{64, 32\} & $0.592\pm0.001$ & 40.8K & $0.655\pm0.001$ & 40.7K\\
    \bottomrule
  \end{tabular}
  \vspace{-0.2cm}  
\end{table*}

\subsection{Model Ablation Study (RQ3, RQ4)}
We conduct the following ablation study on the session-level user engagement prediction task.

\begin{figure}
    \setlength{\abovecaptionskip}{0.cm}
    \subfigure[Ablation study on different model components.]{
        \centering
        \includegraphics[width=1.5in, keepaspectratio]{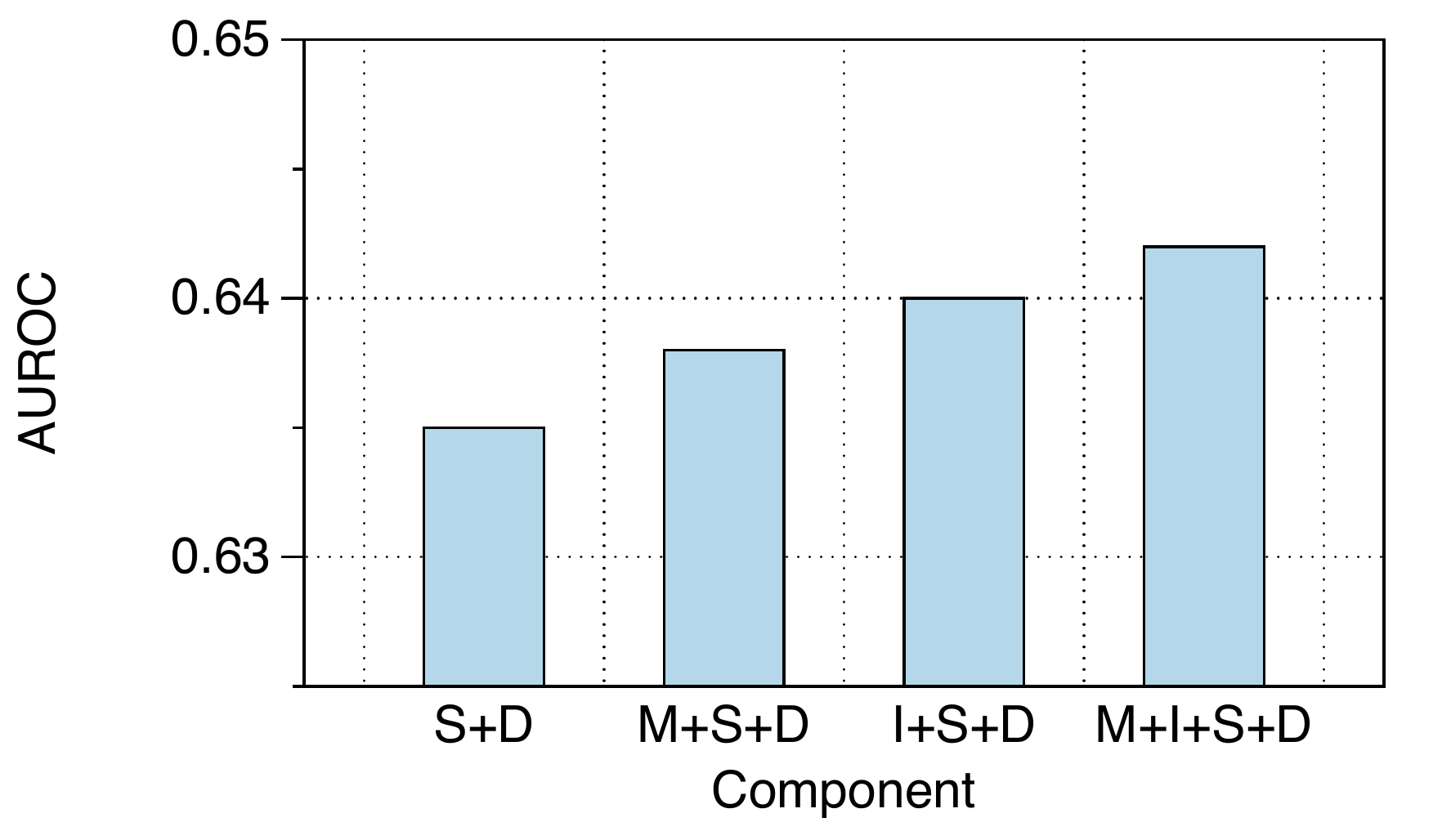}
        \label{Ablation component}
    }
    \subfigure[Ablation study on different parameter adjustment signals of DIGMN.]{
        \centering
	    \includegraphics[width=1.5in, keepaspectratio]{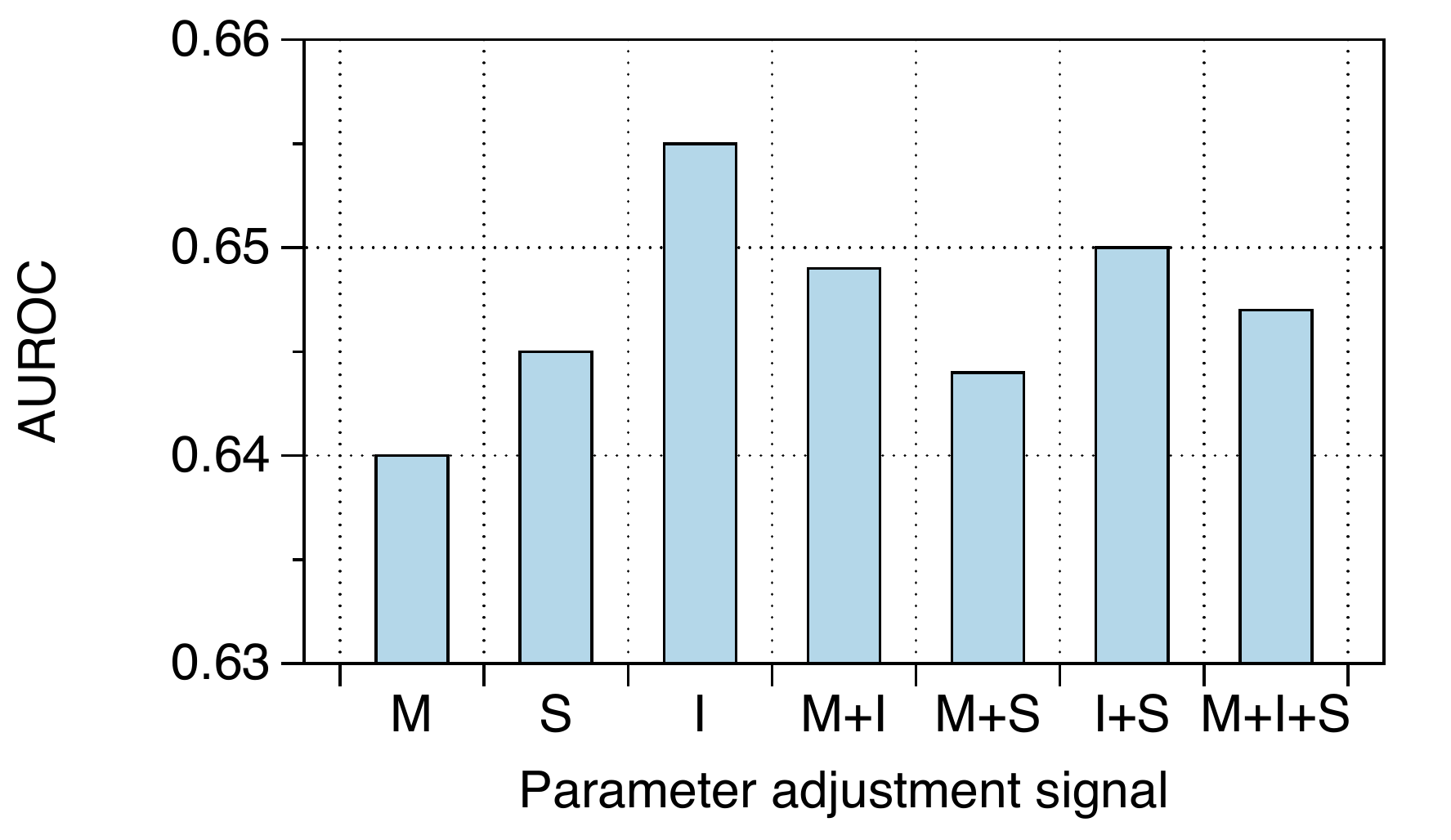}
	    \label{Ablation signal}
    }
    \caption{Ablation studies of components and different parameter adjustment signals. $M, D, S, I$ represent macroscopic features, the latest platform delivery message, session features, and intent features.}
    \vspace{-0.2cm}
\end{figure}

We use an MLP with two hidden layers ($\#$ hidden units are 64 and 32) as the predictor to investigate the effectiveness of various components in our model. As shown in Figure \ref{Ablation component}, we find that simultaneously using the user's macroscopic features, platform delivery message, session features, and intent features can achieve the best results. 

We also study the influence of different adjust signals of DIGMN, as shown in Figure \ref{Ablation signal}. We find that using intent as a parameter adjustment signal performs best, indicating that intent can distinguish different engagement levels well. Introducing too much information to adjust parameters may impair the model classification ability.

\subsection{Hyperparameter Sensitivity (RQ5)}
We conduct the following hyperparameter sensitivity study on the session-level user engagement prediction task.

\begin{figure}[H]
    \setlength{\abovecaptionskip}{0.cm}
    \vspace{-0.4cm}
    \subfigure[Hyperparameter sensitivity of $\bm{a}$.]{
        \centering
        \includegraphics[width=1in, keepaspectratio]{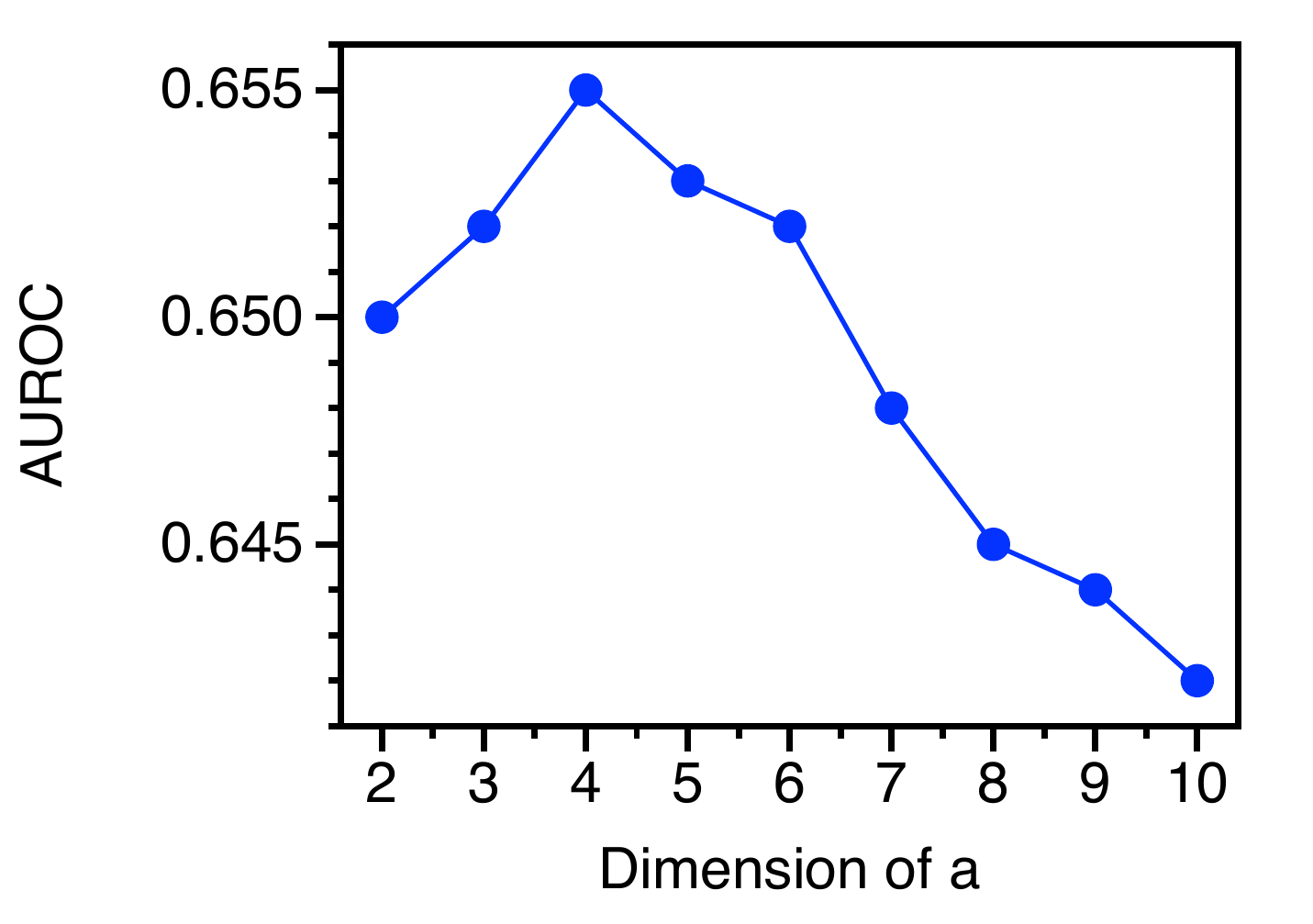}
        \label{Sensitivity_a}
    }
    \subfigure[Hyperparameter sensitivity of $\beta$.]{
        \centering
	    \includegraphics[width=1in, keepaspectratio]{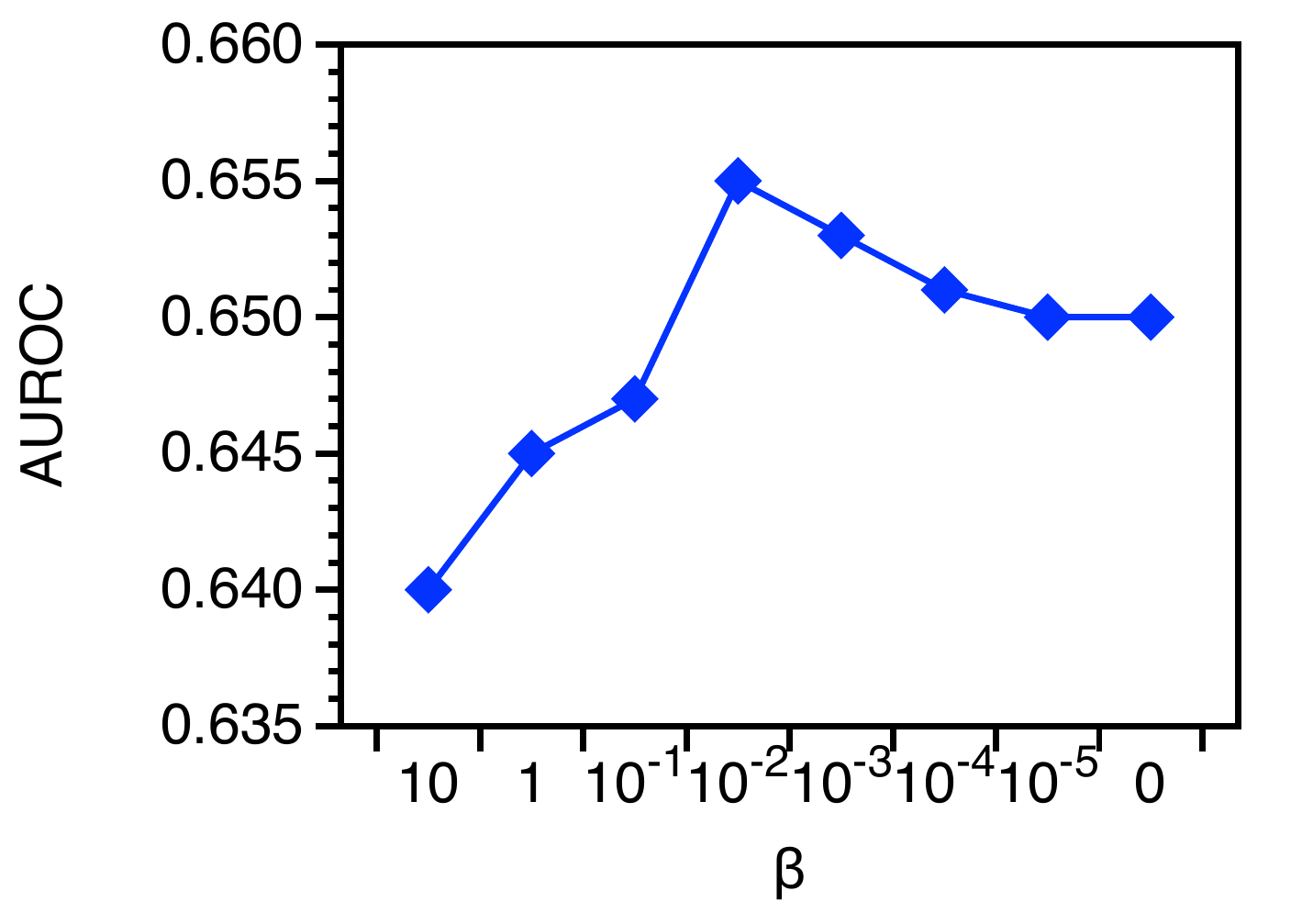}
	    \label{Sensitivity_beta}
    }
    \subfigure[Hyperparameter sensitivity of $L$.]{
        \centering
	    \includegraphics[width=1in, keepaspectratio]{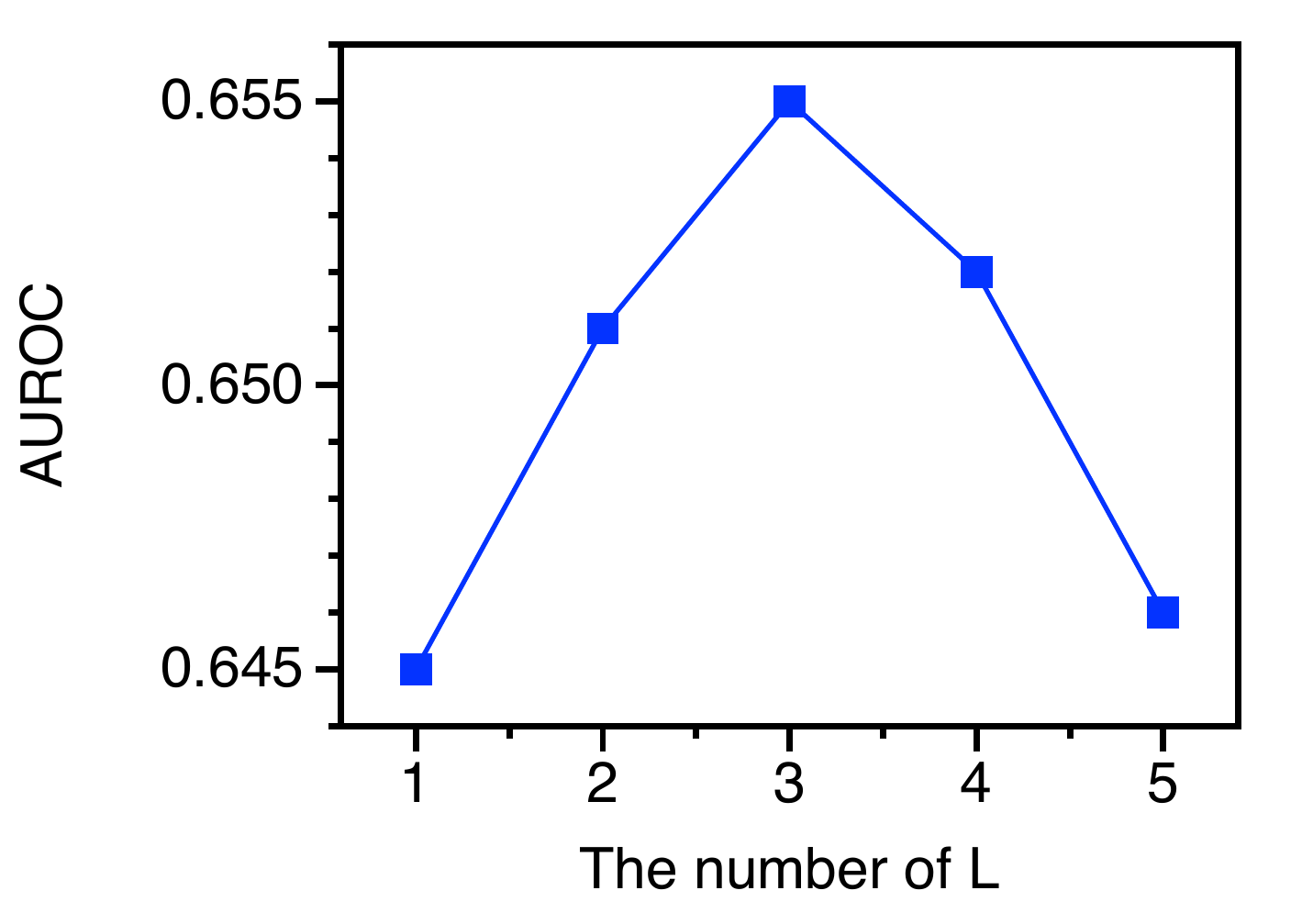}
	    \label{Sensitivity_L}
    }
    \caption{Hyperparameter sensitivity.}
    \vspace{-0.2cm}
\end{figure}

\begin{itemize}[leftmargin=*, itemsep=2pt, topsep=0pt, parsep=0pt]
\item[$\bullet$] \textbf{Hyperparameter sensitivity of $\bm{a}$}: As shown in Figure \ref{Sensitivity_a}, we search for the optimal dimension of $\bm{a}$ between 2 and 10, and find that when the dimension of $\bm{a}$ is too small or too large, it will hurt the model's performance. The possible reason is that when the dimension of $\bm{a}$ is small, the expressive ability of the model is limited and cannot sufficiently capture the diverse engagement patterns of users. However, when the dimension of $\bm{a}$ is large, the number of parameters of the model will rapidly increase, which may cause overfitting and weaken generalization ability. On our dataset, the model performs best when the dimension of $\bm{a}$ is equal to 4.
\item[$\bullet$] \textbf{Hyperparameter sensitivity of $\beta$}: As illustrated in Figure \ref{Sensitivity_beta}, we adopt grid search in $[10, 1, 10^{-1}, 10^{-2}, 10^{-3}, 10^{-4}, 10^{-5}, 0]$, and find that the model performs best when $\beta=10^{-2}$. This indicates that too strong orthogonal constraints on the FC-D layer will damage the classification ability of the model, and adding proper orthogonal constraints to the FC-D layer can improve the model's performance.
\item[$\bullet$] \textbf{Hyperparameter sensitivity of $L$}: As illustrated in Figure \ref{Sensitivity_L}, we vary the number of FC-D layers (i.e., $L$) in the meta predictor from 1 to 5 and find that the model achieves the best performance when $L=3$. The possible reason is that when the number of FC-D layers is small, the learning ability of the model is insufficient, and the model is prone to underfitting. In contrast, when the number of FC-D layers is large, the model is prone to overfitting, which damages the model's generalization ability.
\end{itemize}

\subsection{Further Analysis (RQ6, RQ7)}
In order to compare with other methods for obtaining representations of user intent, we replace the intent inference operation based on cosine similarity computation in the intent evolution layer with a linear mapping (end-to-end learning), which can be formulated as $\bm{i}=\tanh(\bm{W}\bm{\nu}+\bm{b})$, where $\bm{W}\in\mathbb{R}^{10\times7}, \bm{b}\in\mathbb{R}^{7}$ are learnable parameters, $\tanh(\cdot)$ is hyperbolic tangent activation functions used to limit the output value between -1 and 1 (same range as cosine similarity). As shown in Table \ref{LDA_Random}, we can observe that introducing basic prior user intents obtained by LDA to mine user intents in DIGMN performs better than directly learning the implicit intent representation. The possible reason is that the basic prior intents are learned on a larger amount of data, while the end-to-end learning user intent representation usually only uses a part of the data, and the former contains more information.

\begin{table}[H]
  \vspace{-0.2cm} 
  \small
  \caption{Prediction performance of different methods to extract session intents.}
  \label{LDA_Random}
  \begin{tabular}{lc}
    \toprule
    \textbf{Methods to extract session intents} & \textbf{AUROC}\\
    \midrule
    End-to-end learning & $0.644\pm0.001$\\
    Prior intents & $0.655\pm0.001$\\
    \bottomrule
  \end{tabular}
  \vspace{-0.2cm} 
\end{table}

There are usually two ways to implement dynamic parameters: one is to dynamically adjust the model's parameters (which we use in this paper), and the other is to generate the model's parameters directly. The latter method generate transform matrix $\widehat{\bm{W}}^{l}$ of $l$-th FC-D layer, which can be formulated as $\widehat{\bm{W}}^{l}={\rm Reshape}(\bm{W}_{4}({\rm ReLU}(\bm{W}_{3}\bm{\widetilde{i}}+\bm{b}_{3}))+\bm{b}_{4})$, where $\bm{W}_{3}, \bm{W}_{4}, \bm{b}_{3}, \bm{b}_{4}$ are learnable parameters. As shown in Table \ref{ablation_dynamic_fc}, directly generating parameters of FC-D layers usually requires a meta network with more parameters. Meanwhile, during the training process, we find that directly generating model parameters is prone to overfitting which reduces the model's generalization ability.

\begin{table}[H]
  \small
  \caption{Prediction performance of different methods to implement FC-D layer.}
  \label{ablation_dynamic_fc}
  \begin{tabular}{lccc}
    \toprule
    \textbf{Method} & \textbf{\# Parameters} & \textbf{AUROC}\\
    \midrule
    Parameter generation & 240.9K & $0.647\pm0.003$\\
    Parameter adjustment & 40.7K & $0.655\pm0.001$\\
    \bottomrule
  \end{tabular}
  \vspace{-0.2cm} 
\end{table}

\subsection{Visualization (RQ8)}
\begin{figure}[H]
    \setlength{\abovecaptionskip}{0.cm}
    \vspace{-0.4cm}
    \subfigure[The distribution of user dynamic intent in 3-dimensional space.]{
        \centering
        \includegraphics[width=1.5in, keepaspectratio]{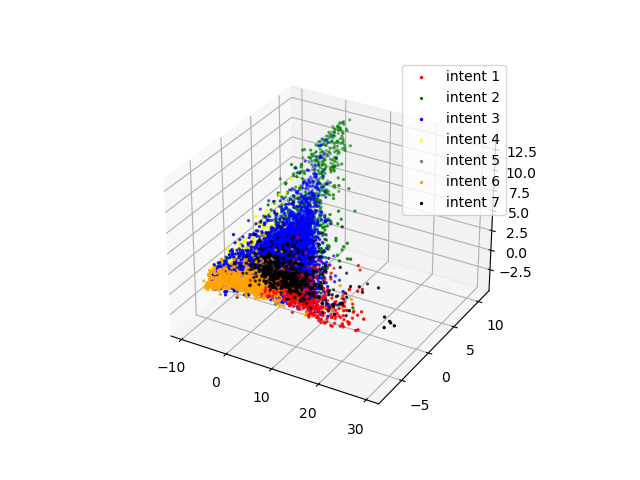}
        \label{3d}
    }
    \subfigure[The distribution of user dynamic intent in 2-dimensional space.]{
        \centering
	    \includegraphics[width=1.5in, keepaspectratio]{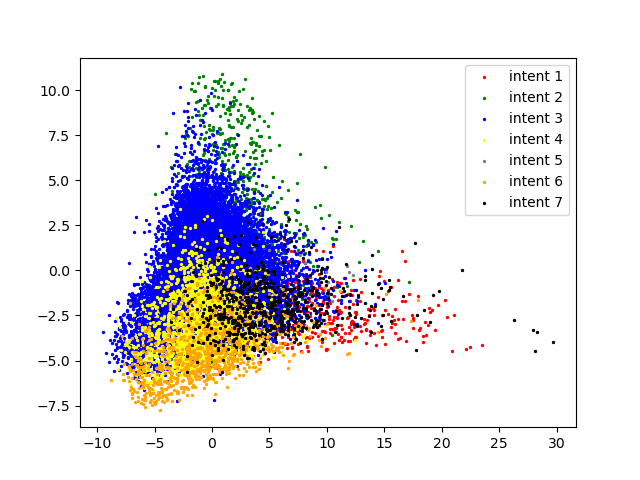}
	    \label{2d}
    }
    \caption{The visualization of dynamic user intent representation. We apply zero-mean normalization to the input high-dimensional vectors before using PCA for dimension reduction. Best viewed in color.}
    \vspace{-0.2cm} 
\end{figure}

We use PCA to reduce the dimensions of dynamic user intent representation from 32 to 3 and 2 in the test dataset of the session-level user engagement forecasting task. The label of each sample is the user's most frequent intent in the past two weeks. Figure \ref{3d} and \ref{2d} show that the intent evolution layer can learn meaningful and discriminative user dynamic intent representation. Subsequent meta-predictor can leverage user dynamic intent representation to adjust network parameters and perform differentiated user engagement prediction.

\section{CONCLUSIONS}
In this work, we use LDA on user session data to mine basic user intents. Meanwhile, we propose a dynamic intent guided meta network (DIGMN) to explicitly model user intent evolution over time and perform differentiated user engagement forecasting. Experiments on the real-world dataset from LinkedIn demonstrate the effectiveness of our method. 

Future work will focus on the interpretability of the method and apply our model to more business scenarios, such as the platform message delivery system. Besides, we will validate our method on more datasets.

\begin{acks}
Thanks to Xu Chen and Yanbin Kang for their insightful discussions.
Thanks to Lu Chen and Yihan Cao for their help in the data process and building the dataset.
Thanks to Yiping Yuan, Huichao Xue, and Yanming Shen for their comprehensive review feedback.
\end{acks}

\bibliographystyle{unsrt}
\bibliography{mybibliography.bib}

\end{document}